\documentclass[letterpaper, 10 pt, conference]{ieeeconf}  

\IEEEoverridecommandlockouts                              

\usepackage[bookmarks=true]{hyperref}
\usepackage{amsmath} 
\usepackage{multicol}
\usepackage{multirow}
\usepackage{amssymb}
\usepackage{array}
\usepackage{empheq}
\usepackage[noadjust]{cite}
\usepackage{color}
\usepackage[export]{adjustbox}
\usepackage[noadjust]{cite}
\usepackage{graphicx} 
\usepackage{cite}
\usepackage{tikz}

\usepackage{comment}

\makeatletter
\newcommand{\thickhline}{%
    \noalign {\ifnum 0=`}\fi \hrule height 1.5pt
    \futurelet \reserved@a \@xhline
}
\newcolumntype{"}{@{\hskip\tabcolsep\vrule width 1pt\hskip\tabcolsep}}
\makeatother

\newcommand\copyrighttext{%
  \footnotesize \textcopyright © 2022 IEEE.  Personal use of this material is permitted.  Permission from IEEE must be obtained for all other uses, in any current or future media, including reprinting/republishing this material for advertising or promotional purposes, creating new collective works, for resale or redistribution to servers or lists, or reuse of any copyrighted component of this work in other works. 
  }
\newcommand\copyrightnotice{%
\begin{tikzpicture}[remember picture,overlay]
\node[anchor=south,yshift=10pt] at (current page.south) {\fbox{\parbox{\dimexpr\textwidth-\fboxsep-\fboxrule\relax}{\copyrighttext}}};
\end{tikzpicture}%
}

\title{\LARGE \bf Efficient Extrinsic Calibration of Multi-Sensor 3D LiDAR Systems for \\ Autonomous Vehicles using Static Objects Information}
\author{
    Brahayam Ponton, Magda Ferri, Lars K\"onig and Marcus Bartels
    \thanks{Authors are with \href{https://www.ibeo-as.com/}{IBEO Automotive Systems GmbH}, Hamburg, Germany}%
    \thanks{This work is part of the project KI-FLEX funded by the German Federal Ministry of Education and Research (BMBF) with project number 16ES1026 as part of the program "AI-based electronic solutions for safe autonomous driving".}
}
\renewcommand{\baselinestretch}{1.0}
\newcommand{\norm}[1]{\lVert#1\rVert}

\newcommand{\ex}{\delta}
\newcommand{\ey}{\varepsilon}
\newcommand{\na}{\mathrm{a}}
\newcommand{\sensors}{\mathcal{S}}
\newcommand{\sensorpair}{(\mathrm{A},\mathrm{B})}
\newcommand{\sensorpairs}{\mathcal{P}}

\newcommand{\sample}{\mathrm{n}^{\sensorpair}}
\newcommand{\samplePlane}{\mathrm{m}^{\sensorpair}}

\newcommand{\x}{\textrm{x}}
\newcommand{\y}{\textrm{y}}
\newcommand{\z}{\textrm{z}}
\newcommand{\yaw}{\theta}
\newcommand{\sinletter}{\psi}
\newcommand{\cosletter}{\zeta}

\renewcommand{\baselinestretch}{0.955}
\allowdisplaybreaks

\begin{document}
\maketitle
\copyrightnotice
\thispagestyle{empty}
\pagestyle{empty}

\begin{abstract}
For an autonomous vehicle, the ability to sense its surroundings and to build an overall representation of the environment by fusing different sensor data streams is fundamental. To this end, the poses of all sensors need to be accurately determined. Traditional calibration methods are based on: 1) using targets specifically designed for calibration purposes in controlled environments, 2) optimizing a quality metric of the point clouds collected while traversing an unknown but static environment, or 3) optimizing the match among per-sensor incremental motion observations along a motion path fulfilling special requirements. In real scenarios, however, the online applicability of these methods can be limited, as they are typically highly dynamic, contain degenerate paths, and require fast computations. In this paper, we propose an approach that tackles some of these challenges by formulating the calibration problem as a joint but structured optimization problem of all sensor calibrations that takes as input a summary of the point cloud information consisting of ground points and pole detections. We demonstrate the efficiency and quality of the results of the proposed approach in a set of experiments with LiDAR simulation and real data from an urban trip.
\end{abstract}

\section{Introduction}

Advanced driver-assistance systems enhance vehicle technology by improving safety and automating driving. For this to happen, an automated vehicle needs to perceive its state and environment, process the available information to come up with a meaningful plan, and use its actuators to put the plan into action. This paradigm is known as "Sense - Think - Act", and highlights that an autonomous vehicle relies on an accurate and robust perception of its environment to properly devise a plan and successfully execute it \cite{SenseThinkActParadigm}.

Consequently, perception systems for automotive applications are designed using several types of sensors such as GNSS data to globally localize the vehicle, IMU systems to gain access to proprioceptive information at high frequencies, and perception sensors (such as LiDARs, cameras, or a combination thereof \cite{Levinson2013AutomaticOC, JointCalibrationMultipleSensors, DAdamo2018RegistrationOT, AutomaticExtrinsicCalibrationGuindel}) to understand the environment. To effectively make use of the sensing redundancies and multi-modality of the sensory system, the different sensors' observations need to be combined continuously in a common coordinate frame using an automated calibration procedure.

Research on the extrinsic calibration of LiDAR sensors has received significant attention, as they offer highly accurate measurements, are not affected by changing light conditions (in contrast to cameras), and the rapid development of their technology over the recent years has led to their widespread use in research and industry applications.

While the simple method of directly measuring the LiDAR position and orientation manually with respect to a reference point on the vehicle could easily provide an accurate measurement of the translation parameters, it would not be able to do the same for the rotational parameters of the calibration \cite{ManualCalibrationKoppanyi}. For this reason, many calibration approaches place visible, unique, and distinctive targets in the scene or use manually labeled control points in the sensor measurements \cite{GeigerMCS12, Unnikrishnan-2005-9235} to accurately estimate all calibration parameters. The drawback of these approaches is that they are time-consuming and require measuring tools, targets, and technical knowledge to perform the calibration procedure. In other words, the accuracy of the results comes at the price of structuring the environment where the calibration will be performed. More recent methods, to be discussed in more detail within Section \ref{sec:related_work}, such as \cite{SensorWithLittleOverlap, ExtCalBrookshire, GloballyOptimalExtCal, ExtCalJeromeMaye,  LostInTranslation, UnsupervisedCalThrun} provide more automated approaches for calibrating sensors using different trade-offs of computational efficiency (i.e. costly computations on point clouds vs. fast computations on egomotions), environmental requirements (such as keeping the environment static during calibration) or trajectory requirements (such as driving along singularity-free trajectories).

The key contributions of this work are:
\begin{enumerate}
    \item A calibration method, that takes as inputs the vehicle egomotion estimates (proprioceptive information) and static objects (perceptual information), and jointly optimizes the extrinsic calibration parameters of all sensors in a structured yet efficient manner.
    \item The proposed method can be used offline (to accurately determine the sensor calibrations) as well as online (to continuously estimate them) thanks to its computational efficiency and capability of working in a natural, dynamic environment (i.e. urban driving with traffic) without any dedicated targets. This brings our algorithm beyond state-of-the-art methods.
    \item An evaluation of the proposed approach with simulation data where ground truth information is available as well as with real data from an urban driving scenario.
\end{enumerate}

The remainder of this work is organized as follows: a brief overview of related work is presented in Section \ref{sec:related_work}. Section \ref{sec:approach} describes in detail the proposed extrinsic calibration algorithm. We show experimental results to evaluate the performance of our approach with simulation as well as real data in Section \ref{sec:experiments} and conclude the paper in Section \ref{sec:conclusion}.

\section{Related Work} \label{sec:related_work}

Many researchers have faced the challenges of estimating the extrinsic calibration parameters, that relate the poses of different sensors, motivated by their application in automotive or robotic domains. In this section, we will discuss contributions relevant to our work.

In \cite{SensorWithLittleOverlap}, a network of laser sensors with little overlaps is cross calibrated using a single dynamic object (i.e. a person, moving through the observed area). In a static environment, sensors that simultaneously detect the only moving object share an overlapping region, and thus, the range measurements to the dynamic object can be used to infer its relative pose. This work is relevant to our approach as it inspires the use of information within overlapping regions of neighboring sensors' FOV (fields of view), to jointly and consistently cross-calibrate them. However, our approach differs from the one presented in \cite{SensorWithLittleOverlap} in that we do not assume a static environment, but a dynamic one where it is our task to figure out which sensor detections are instances of the same object (only seen from different perspectives).

In approaches such as \cite{LostInTranslation, UnsupervisedCalThrun, SheehanCalibration2012}, the extrinsic calibrations of LiDAR sensors with respect to a coordinate frame in the vehicle are estimated based on the optimization of a quality metric over a collection of consecutive point clouds: e.g. \cite{LostInTranslation} optimizes a quadratic entropy metric that quantifies the compactness of the point cloud distribution, and \cite{UnsupervisedCalThrun} minimizes an energy function on the point clouds that penalizes points from being far away from surfaces defined by points from other point clouds. In general terms, they rely on the assumption that points in space tend to lie in contiguous surfaces. These approaches make no assumptions about the environment other than that it is static and rich in 3D features. They differ from ours in that they work directly with point clouds (which renders them accurate but computationally very expensive), while our method relies on using an abstraction that summarizes the point cloud information into higher-level features (such as ground and pole detections). This brings two benefits to the table for our method: 1) it makes our approach computationally efficient as we deal with a lower-dimensional and more structured representation of the environment as input and 2) we overcome the limitation of having a static environment by replacing this requirement with an algorithm capable of detecting static features in a dynamic environment, for which an extensive body of literature exists \cite{AlexPaperOnPolesDetection, ExtractionOfStreetPoleJingmin, AutomaticDetectionClassificationPolesOrdonez}. Note that this detection algorithm \cite{AlexPaperOnPolesDetection} is shared within our architecture with other feature-based components (e.g. localization) and thus does not imply an additional effort for the calibration component.

Another line of calibration methods consists in making use of per-sensor egomotion estimates \cite{ExtCalBrookshire, GloballyOptimalExtCal, ExtCalJeromeMaye}. These methods are capable of recovering the transformation between a pair of sensors mounted rigidly on a moving body using only noisy, per-sensor incremental egomotion observations. \cite{GloballyOptimalExtCal} goes even beyond and provides a method to find globally optimal solutions. These methods require only that the sensors travel together along a motion path that is non-degenerate. This assumption however is limiting, especially when the noise in the measurements is larger than the signal amplitude, as this renders the trajectory degenerate for practical purposes. Our approach combines egomotion information with the information that can be extracted from matching static object detections in overlapping regions of neighboring sensors. This makes it robust against problems where only egomotion information is used at the cost of requiring that sensors do actually have overlapping regions.

Finally, feature-based methods \cite{Zaiter20193DLE, AutomaticExtrinsicRotationalCalibration, Zhou2018AutomaticEC} use environmental features such as lines and planes to compose estimates of the calibration parameters. These methods are close to our approach in terms of the kind of inputs used. However, our method distinguishes itself in that 1) it is also capable of leveraging vehicle egomotion information and 2) it integrates a globally optimal method to efficiently find feasible object correspondences among a set of candidates (constructed based on object detections within the overlapping regions of the fields of view of neighboring sensors).

Now that we have discussed related work in the context of our approach, we will focus on its detailed description.

\section{Extrinsic Calibration Algorithm} \label{sec:approach}

\begin{figure}[b]
	\centering
	\medskip
	\includegraphics[width=\linewidth, trim={4.65cm 5.2cm 6cm 5.1cm}, clip]{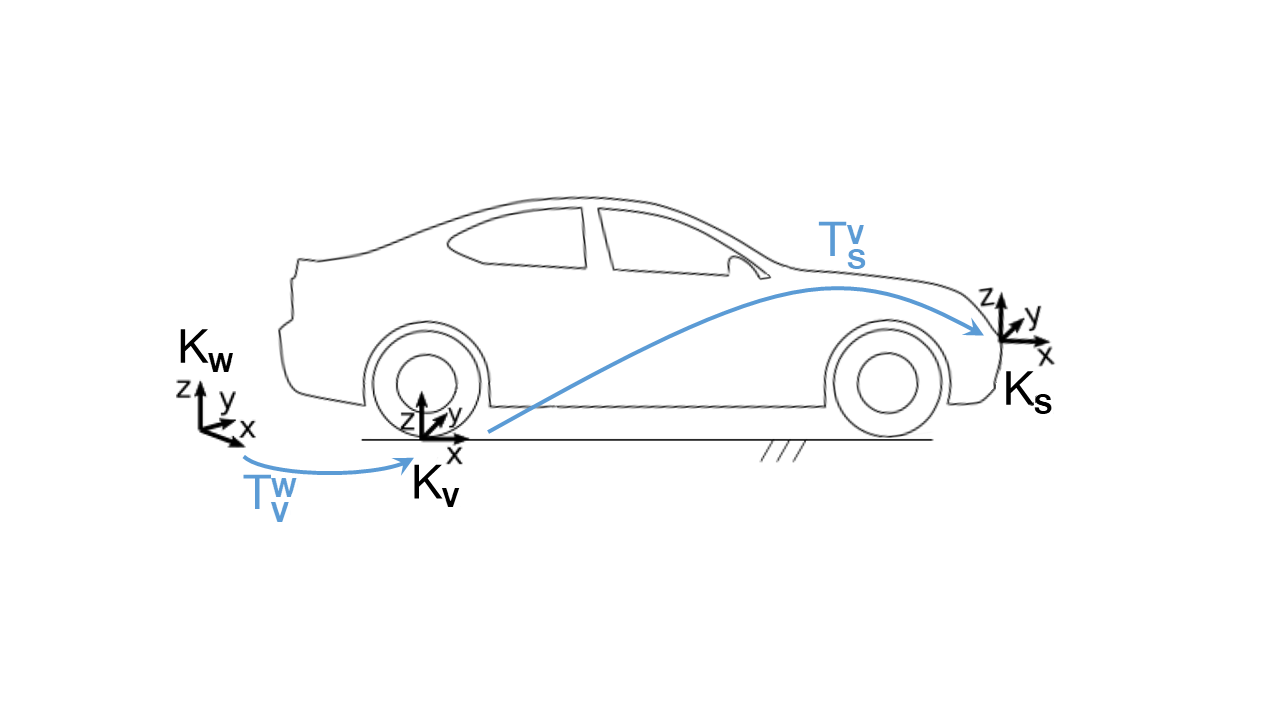}
	\caption[]{\small Definition of relevant coordinate frames in the vehicle.}
	\label{fig:schematic_coords}
\end{figure}

In this section, we describe the offline and online versions of the proposed algorithm for extrinsic calibration. We make this distinction as their pursued objectives are distinct. In the offline version, we start from scratch and our goal is to build an accurate estimate from the available data once. In the online version, we already start from the offline optimized values and the goal is to continuously monitor their correctness, as they might change over time, i.e. due to dynamic effects (e.g. sensor yaw deflection due to the wind), pitch correction of mounting poses to account for changing loads, miscalibrations due to wear and tear, among others.

Before delving into the description of our approach, we introduce some notation in Fig. \ref{fig:schematic_coords}, that defines the coordinate frames to be used throughout this work. $\mathbf{K}_{\mathrm{W}}$ denotes the world reference frame, $\mathbf{K}_{\mathrm{S}}$ the sensor reference frame, and $\mathbf{K}_{\mathrm{V}}$ the vehicle reference frame whose axis system is based on the road (i.e. Z component is normal to the ideal ground plane, and X component points forward). $\mathbf{T}^{\mathrm{W}}_{\mathrm{V}}$ represents the vehicle pose with respect to the world reference frame, and $\mathbf{T}^{\mathrm{V}}_{\mathrm{S}}$ the extrinsic calibration parameters, to be estimated for all sensors. Note that typically an additional coordinate frame fixed to the vehicle body is modeled (that coincides with $\mathbf{K}_{\mathrm{V}}$ in static equilibrium conditions). However, for simplicity we do not show it here, as in the end we are interested in estimating $\mathbf{T}^{\mathrm{V}}_{\mathrm{S}}$.

\subsection{Offline Extrinsic Calibration Method} \label{sec:approach_offline_calibration}

\begin{figure}[t]
	\centering
	\medskip
	\includegraphics[width=\linewidth, trim={2.6cm 7.8cm 11cm 0.8cm}, clip]{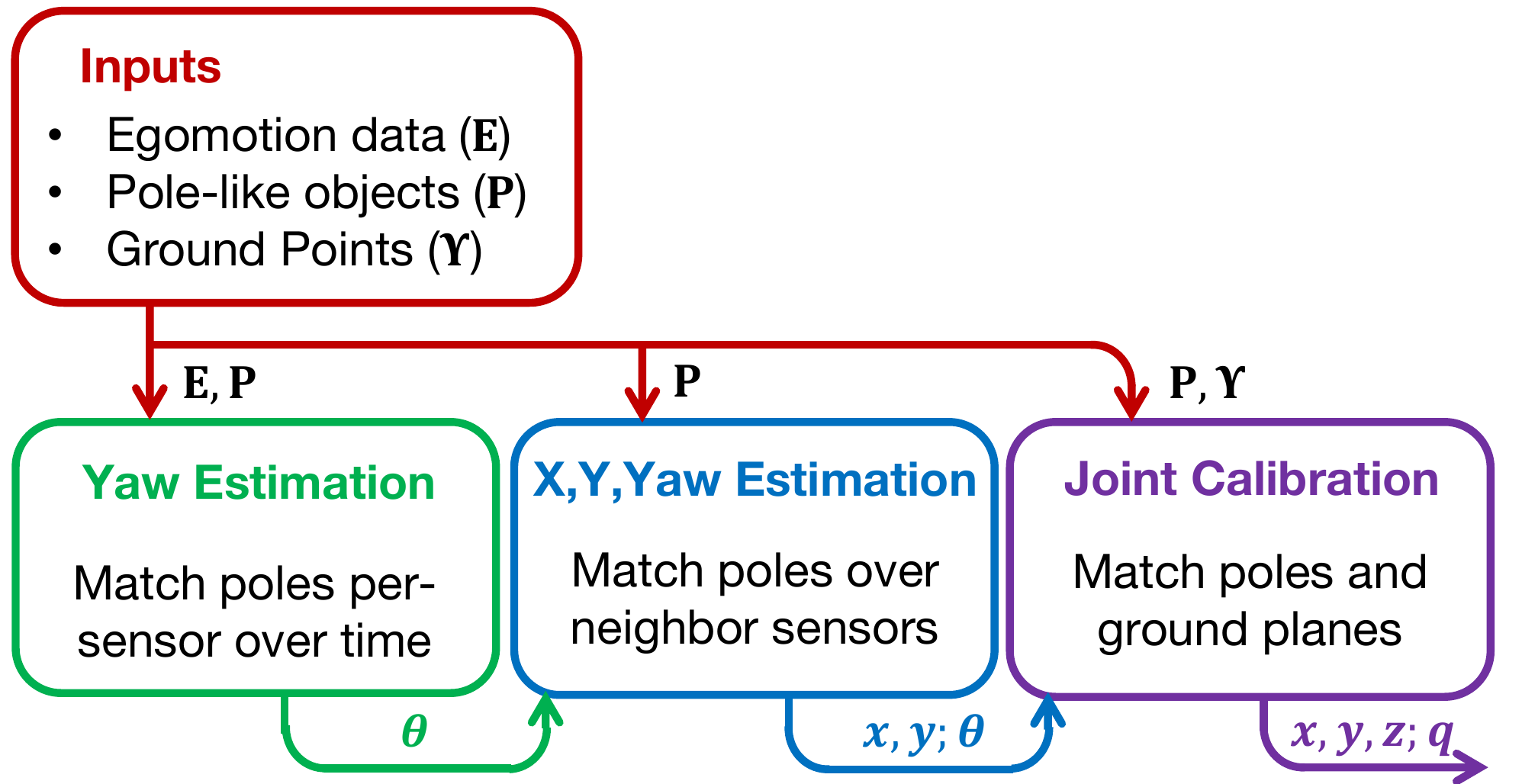}
	\caption[]{\small Block diagram of the offline calibration method. $\x$, $\y$, $\z$ denote the sensor translation, $\yaw$ the yaw angle, and $\mathrm{q}$ its orientation. }
	\label{fig:block_diagram}
\end{figure}

To construct an offline calibration, our algorithm takes as input vehicle egomotion data and per-sensor static information (pole detections and ground points), and outputs a calibration estimate for all sensors by processing the input data through its three core components: 1) an estimator of yaw angles from vehicle egomotion and per-sensor pole detections, 2) an algorithm to match pole pairs for sensors with overlapping regions in their fields of view and 3) an algorithm to jointly calibrate all sensors as shown in Fig. \ref{fig:block_diagram}. In the following, we detail each of these components.

\subsubsection{Yaw estimation} \label{sec:yaw_estimation}
To estimate the sensors' yaw angles in the vehicle frame $\mathbf{K}_{\mathrm{V}}$, we use the relationship between sensor and vehicle incremental egomotions through the extrinsic calibration, under the assumption that this is constant \cite{GloballyOptimalExtCal}. This relationship can be derived as follows:

\begin{subequations}
\begin{align}
\mathbf{T}^{\mathrm{W}}_{\mathrm{S[t]}} &= \mathbf{T}^{\mathrm{W}}_{\mathrm{V[t]}} \mathbf{T}^{\mathrm{V}}_{\mathrm{S}} \label{eq:yaw_rel1}\\
\mathbf{T}^{\mathrm{W}}_{\mathrm{S[t-1]}} \mathbf{T}^{\mathrm{S[t-1]}}_{\mathrm{S[t]}} &= \mathbf{T}^{\mathrm{W}}_{\mathrm{V[t-1]}} \mathbf{T}^{\mathrm{V[t-1]}}_{\mathrm{V[t]}} \mathbf{T}^{\mathrm{V}}_{\mathrm{S}} \label{eq:yaw_rel2}\\
(\mathbf{T}^{\mathrm{W}}_{\mathrm{V[t-1]}})^{-1} \mathbf{T}^{\mathrm{W}}_{\mathrm{S[t-1]}} \mathbf{T}^{\mathrm{S[t-1]}}_{\mathrm{S[t]}} &=  \mathbf{T}^{\mathrm{V[t-1]}}_{\mathrm{V[t]}} \mathbf{T}^{\mathrm{V}}_{\mathrm{S}} \label{eq:yaw_rel3}\\
\mathbf{T}^{\mathrm{V}}_{\mathrm{S}} \mathbf{T}^{\mathrm{S[t-1]}}_{\mathrm{S[t]}} &=  \mathbf{T}^{\mathrm{V[t-1]}}_{\mathrm{V[t]}} \mathbf{T}^{\mathrm{V}}_{\mathrm{S}} \label{eq:yaw_rel4}
\end{align}
\label{eq:yaw_rel}
\end{subequations}

Equation \eqref{eq:yaw_rel1} expresses the relationship between sensor- $\mathbf{T}^{\mathrm{W}}_{\mathrm{S[t]}}$ and vehicle pose $\mathbf{T}^{\mathrm{W}}_{\mathrm{V[t]}}$ via the extrinsic calibration $\mathbf{T}^{\mathrm{V}}_{\mathrm{S}}$ at time $t$. \eqref{eq:yaw_rel2} decomposes the sensor and vehicle poses from $0$ to $t-1$ and from $t-1$ to $t$. A reordering of the terms in \eqref{eq:yaw_rel3} allows us to relate in \eqref{eq:yaw_rel4} incremental sensor egomotions $\mathbf{T}^{\mathrm{S[t-1]}}_{\mathrm{S[t]}}$ to incremental vehicle egomotions $\mathbf{T}^{\mathrm{V[t-1]}}_{\mathrm{V[t]}}$ via the extrinsic calibration parameters $\mathbf{T}^{\mathrm{V}}_{\mathrm{S}}$. Typically, \eqref{eq:yaw_rel4} is used to estimate all calibration parameters when driving along non-degenerate trajectories \cite{ExtCalBrookshire, GloballyOptimalExtCal}.

We use this relationship to relate incremental vehicle egomotions (easily obtained from the vehicle egomotion input) and incremental sensor egomotions (obtained by matching pole detections along time) to compute the sensor yaw angle. To this end, the first step is to find the matching poles from two consecutive sensor observations (pole detections). As the measurements are consecutive in time and come at a high frequency, we can match a pole from observation at time $t-1$ to one at time $t$, if they are the closest in distance and if this distance falls below a maximum threshold.

We define a pole $P$ in terms of the coordinates of its base $P_{\mathrm{base}}$ and top $P_{\mathrm{top}}$ points $\in \mathbb{R}^{3}$. The distance between a pole $P$ and a point $p \in \mathbb{R}^{3}$ is defined as $\mathcal{D}(P,p) = \norm{(p-P_{\mathrm{top}}) \times (p-P_{\mathrm{base}})}_{2} / \norm{P_{\mathrm{top}}-P_{\mathrm{base}}}_{2}$. Note $\mathcal{D}(P,p)$ is independent from the pole length. The distance between a pole $P$ and a pole $Q$ is then defined as $\mathcal{D}(P,Q) = \left(\mathcal{D}(P,Q_{\mathrm{base}})^{2} + \mathcal{D}(P,Q_{\mathrm{top}})^{2}\right)^{1/2}$, in terms of the distances from pole $P$ to the base and top points of pole $Q$.

Once we have computed for each pair of time steps $t-1$ and $t$, a set of matching pole pairs $\mathcal{M}_{t}$ composed of poles $Q_{t-1}$ from time step $t-1$ and poles $P_{t}$ from time step $t$, we can optimize the yaw angles $\theta^{\mathrm{V}}_{\mathrm{S}}$ for each sensor, as follows:

\begin{subequations}
\begin{align}
&\min_{\theta^{\mathrm{V}}_{\mathrm{S}}} \quad \sum_{t=1}^{T} \sum_{Q_{t-1}, P_{t} \in \mathcal{M}_{t}} \mathcal{D} \left( Q_{t-1}, \mathbf{T}^{\mathrm{S[t-1]}}_{\mathrm{S[t]}} P_{t} \right) \label{eq:yaw_opt1}\\
&\mathrm{where } \;\;\; \mathbf{T}^{\mathrm{S[t-1]}}_{\mathrm{S[t]}} = (\mathbf{T}^{\mathrm{V}}_{\mathrm{S}})^{-1} \mathbf{T}^{\mathrm{V[t-1]}}_{\mathrm{V[t]}} \mathbf{T}^{\mathrm{V}}_{\mathrm{S}} \label{eq:yaw_opt2}\\
&\phantom{where } \;\;\; \mathbf{T}^{\mathrm{V}}_{\mathrm{S}} = \begin{bmatrix} \mathbf{R}(0,0,\theta^{\mathrm{V}}_{\mathrm{S}}) & \mathbf{0}^{3 \times 1} \\ 0 & 1 \end{bmatrix} \label{eq:yaw_opt3}
\end{align}
\label{eq:yaw_opt}
\end{subequations}

The above optimization problem \eqref{eq:yaw_opt} estimates per-sensor the yaw angle $\theta^{\mathrm{V}}_{\mathrm{S}}$. This variable is used to compose the calibration guess $\mathbf{T}^{\mathrm{V}}_{\mathrm{S}}$ in \eqref{eq:yaw_opt3} from a zero translation $\mathbf{0}^{3 \times 1}$ and rotation $\mathbf{R}(0,0,\theta^{\mathrm{V}}_{\mathrm{S}}) \in \mathbb{R}^{3 \times 3}$ parameterized by roll, pitch and yaw angles. Here, roll and pitch angles are zero because the sensors are typically mounted close to those values, but if they were e.g. mounted upside down, one could use a better guess of the roll angle, namely $\pi$. The calibration guess $\mathbf{T}^{\mathrm{V}}_{\mathrm{S}}$ is then used in \eqref{eq:yaw_opt2} to compose a guess of the incremental sensor egomotion using the relationship \eqref{eq:yaw_rel4}. Finally, the objective \eqref{eq:yaw_opt1} minimizes for all time steps $t$ and matching pole pairs $\in \mathcal{M}_{t}$, the distance between pole detections in the previous observation $Q_{t-1}$ and pole detections in the current observation mapped to the previous sensor coordinate frame $\mathbf{T}^{\mathrm{S[t-1]}}_{\mathrm{S[t]}} P_{t}$. The latter relationship just rotates and translates the base and top points of poles $P_{t}$ by the guess value of the incremental sensor egomotion $\mathbf{T}^{\mathrm{S[t-1]}}_{\mathrm{S[t]}}$.

We iterate the procedure of matching pole pairs and optimizing the sensor yaw angle until its value converges. In the first attempt to find matching pole pairs, no guess of the incremental sensor egomotion $\mathbf{T}^{\mathrm{S[t-1]}}_{\mathrm{S[t]}}$ is available, but in subsequent attempts, we can use our current guess of it to refine the matching of pole pairs and thus calibration values.

\subsubsection{Estimation of x, y, yaw using overlapping regions} \label{sec:xyyaw_estimation}
The estimated yaw angles give us knowledge about the sensors' distribution around the vehicle. This together with the knowledge about their field of view angles allow us to roughly guess the overlapping angular range between neighboring sensors. For instance, for a pair of sensors $A$ and $B$, whose field of view angles are ${\mathrm{fov}}_{\mathrm{A}}$ and ${\mathrm{fov}}_{\mathrm{B}}$ respectively, if $\frac{1}{2}({\mathrm{fov}}_{\mathrm{A}} + {\mathrm{fov}}_{\mathrm{B}})$ is larger than their yaw angular difference $|\theta_{\mathrm{A}}^{\mathrm{V}} - \theta_{\mathrm{B}}^{\mathrm{V}}|$, then their FOVs overlap and the overlapping angle is the difference $\frac{1}{2}({\mathrm{fov}}_{\mathrm{A}} + {\mathrm{fov}}_{\mathrm{B}}) - |\theta_{\mathrm{A}}^{\mathrm{V}} - \theta_{\mathrm{B}}^{\mathrm{V}}|$. Alternatively a manually defined guess can be used to focus e.g. only on a desired subregion instead of the entire intersection.

The purpose of finding this guess value of the overlapping range between neighboring sensors is to construct candidate matching pole pairs for each pair of neighboring sensors. Note that we call them candidates as we do not know the XY locations of the sensors within the vehicle, and thus cannot guarantee a given pole pair is an actual match just because both lie in the overlapping region of neighboring sensors. The goal of the algorithm from this section is then to find out which pole pairs are feasible as well as the XY locations of the sensors within the vehicle. To this end, we formulate the following mixed-integer optimization problem (MIP):

\begin{subequations}
\begin{align}
    \min_{\substack{\x_{s}, \y_{s}, \yaw_{s} \,\,\forall s \in \sensors\\
        \substack{\ex_{\sensorpair, i}\\
                  \ey_{\sensorpair, i}\\
                  \na_{\sensorpair, i}} \bigg\}
        \substack{\forall \sensorpair \in \sensorpairs, \\ i \in \sample}}} \;\;\; \sum_{\sensorpair \in \sensorpairs} \; \sum_{i \in \sample} ( &|\ex_{\sensorpair, i}| + \nonumber \\[-7ex]
                       &|\ey_{\sensorpair, i}| + \nonumber \\[0ex]
                       &\phantom{a}\na_{\sensorpair, i} ) \label{eq:xyyaw_opt0}
\end{align}
\begin{align}
    &\textrm{linear approximation of yaw rotation (sine $\sinletter$ / cosine $\cosletter$):}\nonumber\\
    \textrm{s.t. } &\sinletter_{\yaw_{s}} = \textrm{m}_{s} \theta_{s} + \textrm{b}_{s} \quad\quad \forall s\in\sensors \label{eq:xyyaw_opt1}\\
    &\cosletter_{\yaw_{s}} = \bar{\textrm{m}}_{s} \theta_{s} + \bar{\textrm{b}}_{s} \label{eq:xyyaw_opt2}\\
    &\mathbf{R}_{\mathrm{yaw}}(\yaw_{s}) = 
        \begin{bmatrix}
            \cosletter_{\yaw_{s}} & -\sinletter_{\yaw_{s}} & 0 \\
            \sinletter_{\yaw_{s}} & \phantom{-}\cosletter_{\yaw_{s}} & 0 \\
            0 & 0 & 1
        \end{bmatrix} \label{eq:xyyaw_opt3}\\
    &\textrm{definition of sensor calibration:}\nonumber\\
    &\mathbf{T}^{\mathrm{V}}_{\mathrm{s}} = \begin{bmatrix} \mathbf{R}_{\mathrm{yaw}}(\theta_{s}) \mathbf{R}_{\mathrm{pitch}}(0) \mathbf{R}_{\mathrm{roll}}(0) & \begin{bmatrix} \x_{s} \\ \y_{s} \\ 0 \end{bmatrix} \\ 0 & 1 \end{bmatrix} \label{eq:xyyaw_opt4} \\
    &\textrm{bounds on sensor variables:}\nonumber\\
    &\begin{bmatrix} -\frac{\ell}{2} + o_{\x} \\ -\frac{\omega}{2} + o_{\y} \\ \yaw^{*}_{s}-\gamma \end{bmatrix} \leq
    \begin{bmatrix} \x_{s} \\ \y_{s} \\ \yaw_{s} \end{bmatrix} \leq
    \begin{bmatrix} \frac{\ell}{2} + o_{\x} \\ \frac{\omega}{2} + o_{\y} \\ \yaw^{*}_{s}+\gamma \end{bmatrix} \label{eq:xyyaw_opt5} \\[2ex]
    &\textrm{selection of feasible pole pairs from candidates:}\nonumber\\
    &\na_{\sensorpair, i} \in \{0, 1\} \quad\quad\;\; \forall \sensorpair \in \sensorpairs, i \in \sample \label{eq:xyyaw_opt6} \\
    &\na_{\sensorpair, i} \Rightarrow
        \begin{cases}
	        \begin{aligned}
	        \begin{bmatrix} \ex_{\sensorpair, i} \\ \ey_{\sensorpair, i} \end{bmatrix} =
	        \begin{matrix} \mathbf{T}^{\mathrm{V}}_{\mathrm{A}} (P_{\mathrm{base}})_{A, i} \;- \\ \mathbf{T}^{\mathrm{V}}_{\mathrm{B}} (P_{\mathrm{base}})_{B, i} \end{matrix} \\[0.5ex]
	        \end{aligned} & \textrm{, if 0} \\
	        \begin{bmatrix} \ex_{\sensorpair, i} \\ \ey_{\sensorpair, i} \end{bmatrix} = 0 & \textrm{, if 1}
        \end{cases} \label{eq:xyyaw_opt7} \\
    &\textrm{maximum allowed error induced  by a pole pair:}\nonumber\\
    &|\ex_{\sensorpair, i}| + |\ey_{\sensorpair, i}| \leq \lambda \label{eq:xyyaw_opt8}
\end{align}
\label{eq:xyyaw_opt}
\end{subequations}

\noindent This MIP optimizes the calibration variables $\x_{s}, \y_{s}, \yaw_{s}$ denoting the XY translations and yaw orientations (expressed in vehicle frame  $\mathbf{K}_{\mathrm{V}}$) for all sensors $s\in\sensors$. It also includes helper variables such as binary decisions $\na_{\sensorpair, i}$ (that define if a candidate pole pair is an actual match [0] or not [1]) and error variables $\ex_{\sensorpair, i}$, $\ey_{\sensorpair, i}$ (that define the amount of error induced by a pole pair when it is considered a match). Helper variables are indexed for all sensor pairs in the set of neighboring sensors $\forall \sensorpair \in \sensorpairs$ and every candidate pole pair for a given sensor pair $i \in \sample$.

The problem includes the following constraints: \eqref{eq:xyyaw_opt1}-\eqref{eq:xyyaw_opt2} define a linear approximation of sine and cosine of yaw angles $\theta_{s}$ at the values optimized in the previous step $\theta^{*}_{s}$. $\mathrm{m}_{s} = \mathrm{cos}(\theta^{*}_{s})$ denotes the slope of the sine linear approximation, and $\textrm{b}_{s} =\mathrm{sin}(\theta^{*}_{s}) - \mathrm{m}_{s} \theta_{s}$ its intercept. $\bar{\mathrm{m}}_{s} = -\mathrm{sin}(\theta^{*}_{s})$ denotes the slope of the cosine linear approximation, and $\bar{\mathrm{b}}_{s} = \mathrm{cos}(\theta^{*}_{s}) - \bar{\mathrm{m}}_{s} \theta_{s}$ its intercept. \eqref{eq:xyyaw_opt3} denotes the linear approximation of the yaw rotation matrix based on \eqref{eq:xyyaw_opt1}-\eqref{eq:xyyaw_opt2}. \eqref{eq:xyyaw_opt4} defines the sensor calibration $\mathbf{T}^{\mathrm{V}}_{\mathrm{s}}$ with all parameters optimized in this step. Similarly, as in the previous yaw estimator, this method accepts a different initial guess of roll and pitch orientations. \eqref{eq:xyyaw_opt5} defines limits for all calibration variables, i.e. yaw angles $\theta_{s}$ are constrained to lie close to its initial guess $\theta^{*}_{s}$ to maintain the validity of the linear approximations, $\x_{s}$ and $\y_{s}$ sensor translations are constrained to lie within the boundaries of the vehicle of length $\ell$ and width $\omega$. The constants $o_{\x}$ and $o_{\y}$ are offsets that allow to account for the fact that the vehicle reference frame $\mathbf{K}_{\mathrm{V}}$ is not located at the center of the vehicle but in the middle of the rear axle. The algorithm assumes that vehicle dimensions ($\ell$, $\omega$) and offsets ($o_{\x}$, $o_{\y}$) can be measured. All constraints \eqref{eq:xyyaw_opt1}-\eqref{eq:xyyaw_opt5} are defined for all sensors $\forall s \in \sensors$.

The last three constraints are defined for all neighboring sensor pairs $\forall \sensorpair \in \sensorpairs$ and all candidate pole pairs for a given sensor pair $i \in \sample$. \eqref{eq:xyyaw_opt6} defines that variables $\na_{\sensorpair, i}$ are binary and \eqref{eq:xyyaw_opt7} defines the meaning of the error variables ($\ex_{\sensorpair, i}$, $\ey_{\sensorpair, i}$) depending on the value taken by the binary variables $\na_{\sensorpair, i}$. When a candidate pole pair is not selected as feasible ($\na_{\sensorpair, i} = 1$), the error variables are defined as zero, so that they do not affect the cost. On the other hand, when a candidate pole pair is selected as feasible ($\na_{\sensorpair, i} = 0$), the error variables are defined as the difference (XY components) of the poles from the candidate pole pair expressed in the vehicle reference frame $\mathbf{K}_{\mathrm{V}}$. \eqref{eq:xyyaw_opt8} defines a maximum threshold $\lambda$ for the matching error between the poles of the candidate pole pair.

Finally, note that this MIP minimizes the absolute value norm (1-norm) of the error variables ($\ex_{\sensorpair, i}$,  $\ey_{\sensorpair, i}$) for the pole pairs selected as feasible matches, as they are zero otherwise. At the same time, we minimize the number of candidate pole pairs not selected as feasible, as it increases the objective (as $\na_{\sensorpair, i}$ takes the value of 1 in this case). This avoids trivial solutions where no pole pair is selected. In addition, a regularization of yaw values $\yaw_{s}$ to its initial values $\yaw^{*}_{s}$ keeps this solution close to the one from the previous step.

The outputs of this component (used as input for the next one) are the currently optimized values for XY translations and yaw orientations for all sensors, as well as the set of feasible pole pairs for all neighboring sensor pairs (those for which $\na_{\sensorpair, i}$ equals zero). Note that this problem is efficiently solvable (despite being a mixed-integer one) as internally it just needs to solve linear problem descriptions. 

\subsubsection{Joint cross-sensor calibration} \label{sec:cross_estimation}

Within this last calibration step, we will finely tune all of the calibration parameters using the results from the previous steps as well as the remaining input information (ground points). In this step, we solve the following optimization problem:

\begin{align}
    \min_{\mathbf{T}^{\mathrm{V}}_{s} \forall s\in\sensors} \quad& \sum_{s\in\sensors} \underbrace{\norm{\mathbf{T}^{\mathrm{V}}_{s} \ominus (\mathbf{T}^{\mathrm{V}}_{s})^{*}}^{2}}_{\textrm{Regularization term}} + \nonumber\\
    &\underbrace{\sum_{\sensorpair \in \sensorpairs} \sum_{\substack{i \in \sample \\ \forall \na_{\sensorpair, i} = 0}} \mathcal{D}(\mathbf{T}^{\mathrm{V}}_{\mathrm{A}} P_{A,i}, \mathbf{T}^{\mathrm{V}}_{\mathrm{B}} P_{B,i})}_{\textrm{Pole-matching cost term}} + \nonumber\\
    &\underbrace{\sum_{\sensorpair \in \sensorpairs} \; \sum_{j \in \samplePlane} \mathcal{D}_{\mathrm{pl}}(\mathbf{T}^{\mathrm{V}}_{\mathrm{A}} Pl_{A,j}, \mathbf{T}^{\mathrm{V}}_{\mathrm{B}} Pl_{B,j})}_{\textrm{Plane-matching cost term}} +\nonumber\\[0ex]
    &\underbrace{\mathcal{D}_{\mathrm{ang}}(\mathbf{T}^{\mathrm{V}}_{\mathrm{A}} Pl_{A,j}, \mathcal{G}) + \mathcal{D}_{\mathrm{ang}}(\mathbf{T}^{\mathrm{V}}_{\mathrm{B}} Pl_{B,j}, \mathcal{G})}_{\textrm{Angular difference to ideal ground plane}}
\label{eq:cross_opt}
\end{align}

This problem optimizes all sensors' calibrations $\mathbf{T}^{\mathrm{V}}_{s}\;\forall s\in\sensors$, composed of 3d translations and 4d quaternions. The objective regularizes the XY translational components and yaw orientation of all sensors to the values found in the previous step $(\mathbf{T}^{\mathrm{V}}_{s})^{*}$. The symbol $\ominus$ denotes a logarithmic map for each of the components (translation and orientation) separately \cite{LieAlgebraReference}. We use quaternions as they offer a compact singularity-free representation of orientations, which makes it possible to efficiently optimize them using the properties of its Lie-algebra. In particular, the logarithmic map is an isomorphism, meaning that angular differences between two orientations are the same independent from where they are measured. It rewards good matching of all feasible pole pairs for all neighboring sensors. All feasible pole pairs (composed of a pole in sensor $A$ coordinates $P_{A,i}$, and one in sensor $B$ coordinates $P_{B,i}$) are mapped through the corresponding calibrations to the vehicle frame $\mathbf{K}_{\mathrm{V}}$ as $\mathbf{T}^{\mathrm{V}}_{\mathrm{A}} P_{A,i}$ and $\mathbf{T}^{\mathrm{V}}_{\mathrm{B}} P_{B,i}$, where their distance $\mathcal{D}$ is then minimized.

The third term in the optimization objective is a running sum over all sensor pairs $\sensorpair \in \sensorpairs$ and all plane pairs per sensor pair $j \in \samplePlane$. By plane pairs, we mean the planes fitted to the ground points within the overlapping regions of neighboring sensors, when they satisfy certain validity conditions (i.e. there is a minimum number of points to build the plane, their relative orientation is below a desired threshold, among others). All plane pairs (composed of a plane in sensor $A$ coordinates $Pl_{A,j}$, and a plane in sensor $B$ coordinates $Pl_{B,j}$) are mapped to the vehicle frame $\mathbf{K}_{\mathrm{V}}$ as $\mathbf{T}^{\mathrm{V}}_{\mathrm{A}} Pl_{A,j}$ and $\mathbf{T}^{\mathrm{V}}_{\mathrm{B}} Pl_{B,j}$, where their distance $\mathcal{D}_{\mathrm{pl}}$ can be minimized. $\mathcal{D}_{\mathrm{pl}}$ is defined as the sum of the distances between three non-colinear points from one plane and the other. A plane $Pl_{A}$ is defined by a plane point ${\mathrm{pnt}}_{A}$ (centroid of ground points within the overlapping region) and a normal vector $\eta_{A}$ (computed using a spectral decomposition of the at zero centered matrix of ground points). The three non-colinear points are composed of the centroid and the centroid moved along the plane tangent directions (from previous SVD decomposition). Thus, the distance between plane $Pl_{A}$ and point $p\in\mathbb{R}^{3}$ is $(\eta_{A}/ \norm{\eta_{A}}_{2}) \cdot (p - {\mathrm{pnt}}_{A})$. As the fourth term, the objective regularizes ground planes orientation to the ideal ground plane $\mathcal{G}$ (assumed to be locally flat in urban scenarios). This happens by minimizing the angle $\mathcal{D}_{\mathrm{ang}}$ between the planes in the vehicle frame ($\mathbf{T}^{\mathrm{V}}_{\mathrm{A}} Pl_{A,j}$, $\mathbf{T}^{\mathrm{V}}_{\mathrm{B}} Pl_{B,j}$) and the ideal ground plane $\mathcal{G}$. This angular distance is simply the scalar product of the normalized plane normals.

Note that in an urban environment, not all sensors see ground points all the time, sometimes they see sidewalks or even the sides of other cars. This is the reason why in the above approach we do not independently compute each sensor's height from the ground. Instead, we calibrate them relative to each other by matching planes fitted to ground points over the overlapping regions of neighboring sensors (which are very likely to see the same points and thus fit a similar plane). However, in the end, as we need a calibration with respect to $\mathbf{K}_{\mathrm{V}}$, which lies on the ground, we need to compute the absolute height for one of the sensors (e.g. one looking to the front). With this absolute height estimate and the relative ones (we have previously optimized), we can estimate all sensor heights with respect to $\mathbf{K}_{\mathrm{V}}$.

Note, that we normalize in all cost functions the contributions of all sensors and neighboring sensor pairs in the vehicle such that they play a similar role in the overall cost.

To summarize, in this section we presented an approach to jointly calibrate all sensors of a vehicle by first using information along the time dimension, namely vehicle egomotion estimates and per-sensor matching poles over time, to compute the yaw angles at which the sensors are mounted. Then in a second step, we exploit information along the space dimension, i.e. we find objects that are simultaneously seen by neighboring sensors and thus provide us information about their relative locations. Finally, we build an overall calibration estimate for all sensors by cross-calibrating them using all the information available so far.

\subsection{Online Extrinsic Calibration Method} \label{sec:approach_online_calibration}

In this section, we provide a brief overview of how our algorithm is adapted to run online. In principle, it makes use of the previously presented ideas with minor simplifications for efficiency reasons. It is composed of 3 components: the yaw estimator \eqref{eq:yaw_opt}, an estimator for roll, pitch, and sensor heights based on \eqref{eq:cross_opt}, and an algorithm for fast alignment of all sensors based on \eqref{eq:xyyaw_opt} but without binary variables. In the following, we describe these algorithms in the order they are used and explain how they differ from the previous ones.

\subsubsection{Yaw estimation} \label{sec:yaw_estimation_online}
The yaw estimator for online calibration is similar to the one from section \ref{sec:yaw_estimation}. The major difference is that it starts with an already good calibration guess, which is leveraged for the selection of pole matches at consecutive time steps, as well as within the optimization problem \eqref{eq:yaw_opt}, where a guess for all the calibration parameters is available to construct \eqref{eq:yaw_opt3}. In other words, we maintain a current estimate for all extrinsic calibrations that can be used within this estimator, which in turn only updates the current estimate of yaw angles for all sensors independently.

\subsubsection{Roll, pitch and heights estimator} \label{sec:rph_estimation_online}
We then proceed to make updates to the current values of roll, pitch, and relative heights of all sensors using an optimization problem similar to \eqref{eq:cross_opt}. The reason why the algorithm from section \ref{sec:cross_estimation} is not directly used is twofold: first, it is computationally expensive as it optimizes all sensor calibrations, while in this case we only make updates to half of the parameters. The second and most important reason is that \eqref{eq:cross_opt} requires as input a set of feasible pole pairs for all sensor pairs, which could take time to accumulate before we can actually use it to make updates to our calibration parameters (thus rendering our approach slow when reacting to miscalibrations). To avoid this, in the online version we use the algorithm from section \ref{sec:cross_estimation} but without considering the cost and input related to the set of feasible pole pairs for all sensor pairs, i.e. we ignore the pole matching cost $\mathcal{D}(\mathbf{T}^{\mathrm{V}}_{\mathrm{A}} P_{A,i}, \mathbf{T}^{\mathrm{V}}_{\mathrm{B}} P_{B,i}) \, \forall \sensorpair \in \sensorpairs, i \in \sample$ from \eqref{eq:cross_opt}. Similarly, as in the yaw estimator, we regularize all calibration values to the current ones.

\subsubsection{Estimation of x, y, yaw using overlapping regions} \label{sec:xyyaw_estimation_online}
This estimator achieves an overall alignment of all sensors similarly to \eqref{eq:xyyaw_opt}, but with the following modifications: First, we can exploit the estimate of the XY sensor translations to find feasible pole pairs for neighboring sensors based only on distance. This frees us from having to decide which candidate pole pairs are actually feasible, and thus we can safely remove from the optimization all binary variables $\na_{\sensorpair,i}$ $\forall \sensorpair \in \sensorpairs, i \in \sample$ and replace the constraints \eqref{eq:xyyaw_opt6}-\eqref{eq:xyyaw_opt8} by a single one that defines the values of the error variables (similar to \eqref{eq:xyyaw_opt7} when $\na_{\sensorpair,i} = 0$). \eqref{eq:xyyaw_opt4} can also use all values of the sensor calibrations (e.g. for pitch, roll, and sensor heights). Finally, the cost in \eqref{eq:xyyaw_opt0} discards penalties on binary decision variables, updates the penalties over error variables from 1-norm to 2-norm to improve convergence and more highly penalize large mismatches of pole base points (seen from different sensors), and regularizes yaw angles and XY translations to their current values.

\section{Experimental Results} \label{sec:experiments}

In this section, we will present a qualitative and quantitative evaluation of our calibration approach using simulation data (for which ground truth data is available) as well as real data from an urban trip in one of our test vehicles. The vehicle is equipped with 8 latest-generation fully solid-state ibeoNEXT LiDAR sensors \cite{IbeoSensors} with 60° FOV, mounted such that they together provide a 360° FOV. The trip took place in an unmodified urban environment of Berlin-Reinickendorf.

\subsection{Evaluation of Offline Calibration Approach}

In this section, we will start with a qualitative evaluation of our method for offline calibration (using a 30sec snippet of real data during an urban trip) to exemplify the way it works and the quality of the resulting calibrations.

\begin{figure}[t]
	\centering
	\medskip
	\includegraphics[width=\linewidth, trim={0cm 0cm 0cm 0.5cm}, clip]{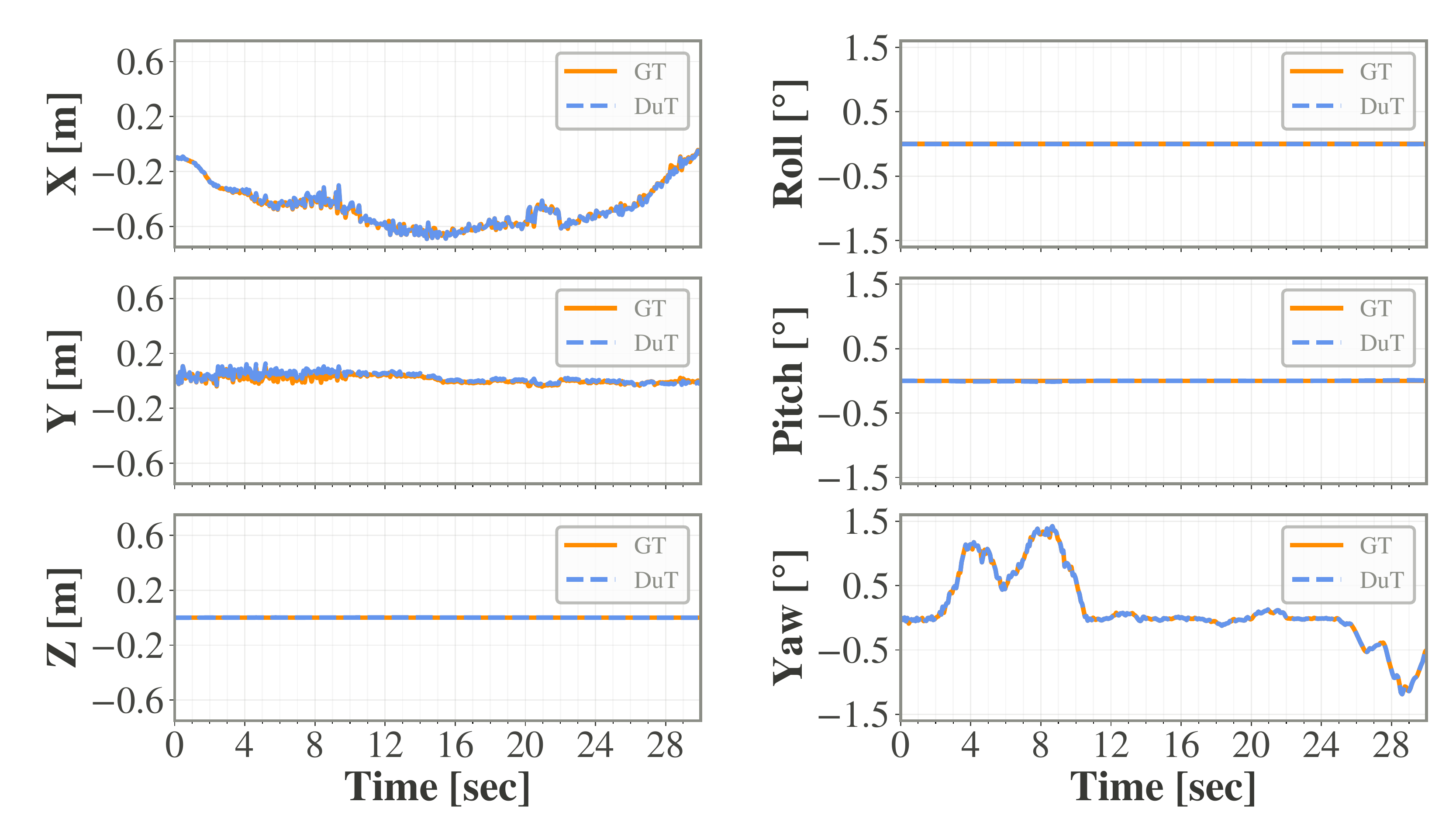} \\[0.3em]
	\includegraphics[width=0.97\linewidth, trim={0cm 6cm 6cm 6cm}, clip]{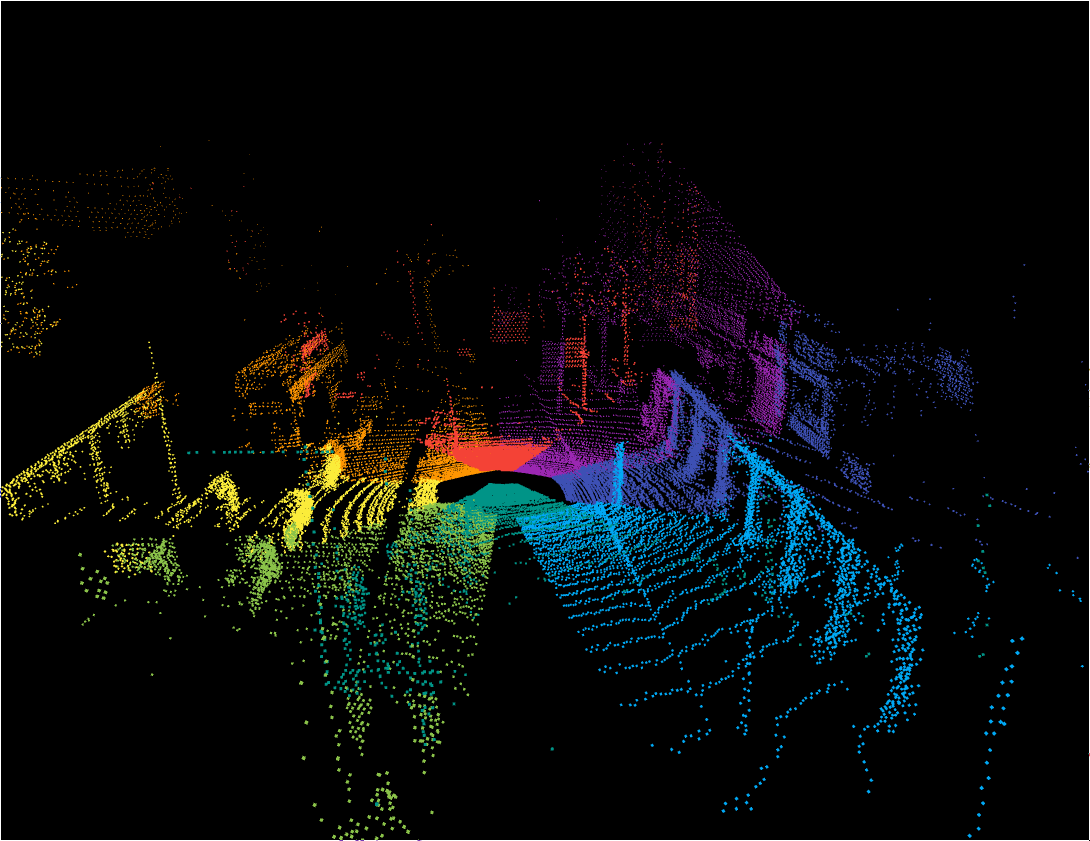}
	\caption[]{\small \textit{Top}: Comparison of incremental sensor egomotions (obtained from recorded egomotion data) predicted with yaw calibrations only (DuT) and with ground truth calibrations (GT) obtained using a static calibration procedure. \textit{Bottom}: Alignment of point clouds after yaw calibration only (algorithm from section \ref{sec:yaw_estimation}).}
	\label{fig:odometries_comparison}
\end{figure}

First of all, we look into the intermediate results obtained by the first part of the algorithm, which estimates the yaw angles (algorithm from section \ref{sec:yaw_estimation}). Remember that this algorithm predicts incremental sensor egomotions \eqref{eq:yaw_opt2} and a calibration estimate \eqref{eq:yaw_opt3} to match static object detections between consecutive time steps. Thus, it is interesting to see the predictions for $\mathbf{T}^{\mathrm{S[t-1]}}_{\mathrm{S[t]}}$ using ground truth calibrations (including all parameters) and the calibrations using only the currently optimized yaw angles. Fig. \ref{fig:odometries_comparison} top-view shows exactly this information for one of the sensors mounted on the vehicle and one can easily identify two aspects: 1) both trajectories match closely, which means that estimating the yaw angle extracts almost all of the information within this data, which explains why extracting the rest of parameters from it is inaccurate, and 2) vertical motion changes, roll and pitch angular changes are almost zero, which in practical terms can be interpreted as the trajectory being numerically degenerate (not strong enough incremental egomotion signals to infer all calibrations). These observations were key for designing the calibration algorithm as shown in section \ref{sec:approach_offline_calibration}. Fig. \ref{fig:odometries_comparison} bottom-view shows the yaw alignment of the point clouds corresponding to 8 sensors mounted on the test vehicle. Notice how this first step correctly captures the distribution of all the sensors around the vehicle.

\begin{figure}[t]
	\centering
	\medskip
	\includegraphics[width=\linewidth, trim={2cm 0cm 4.5cm 2cm}, clip]{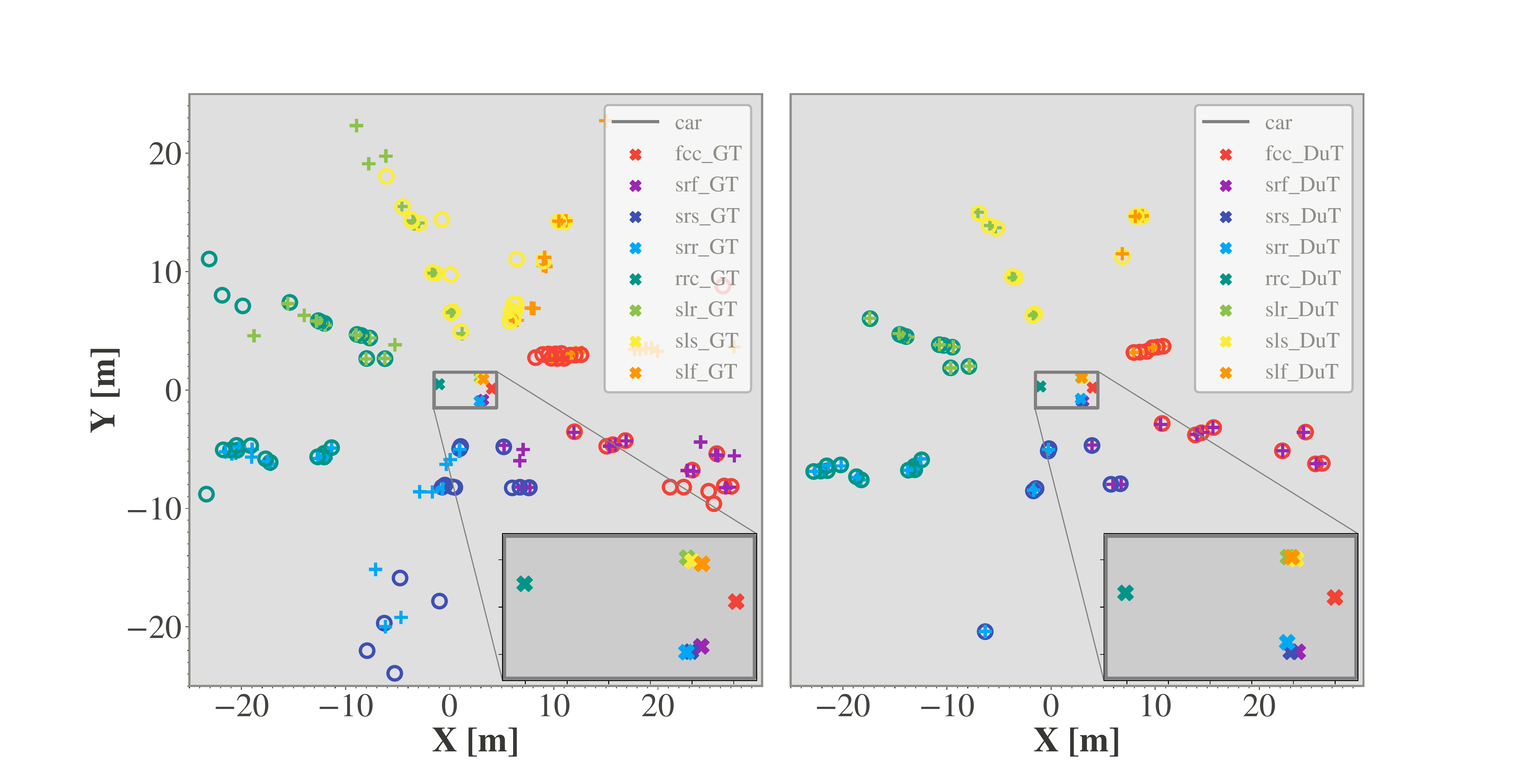}
	\includegraphics[width=0.98\linewidth, trim={0cm 6cm 6cm 6cm}, clip]{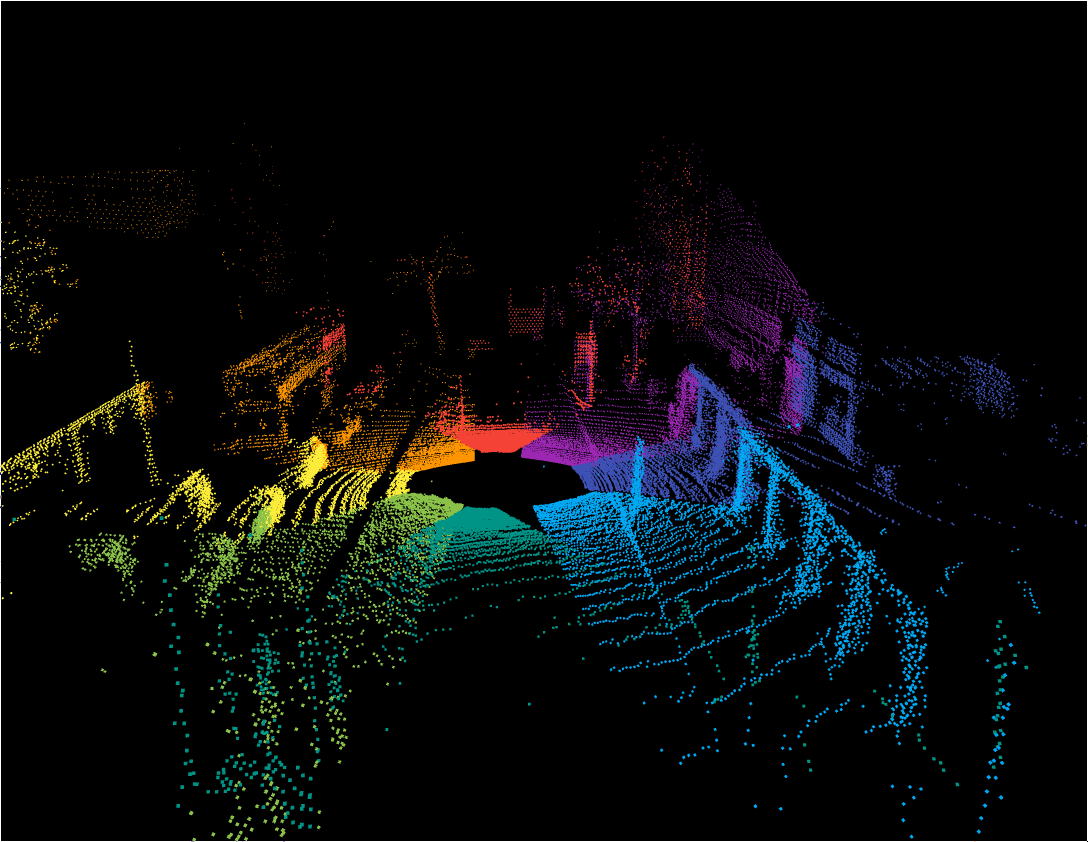}
	\caption[]{\small \textit{Top}: Selection of feasible pole pairs (right), from all candidates (left) based on algorithm from section \ref{sec:xyyaw_estimation}. \textit{Bottom}: Alignment of point clouds after second calibration step. At this point, only $\x$, $\y$ and $\yaw$ components are calibrated.}
	\label{fig:pole_pairs_selection}
\end{figure}

The second part of the calibration algorithm (described in section \ref{sec:xyyaw_estimation}) optimizes the XY translations of the sensors within the vehicle dimensions and slightly updates yaw orientations to achieve a better overall alignment. The criteria under which they are selected is that they should maximize the number of pole pairs selected as feasible, while at the same time minimizing the matching error between the selected pole pairs. On the top of Fig. \ref{fig:pole_pairs_selection}, we show side by side a subset of the candidate pole pairs seen throughout the trip (to the left) and the pole pairs selected as feasible ones (to the right). To the left, we show the 8 sensors at their ground truth calibration values and outside the vehicle, we show the poles seen by each sensor with the corresponding sensor color. In this view, it is easy to see that some candidate pole pairs match (poles detected by two sensors) and others do not. The task of our algorithm is to find out what are the sensor translations that maximize the number of feasible pole pairs and minimize their matching error. The results of the algorithm at selecting feasible pole pairs and estimating XY sensor translations are visible in Fig. \ref{fig:pole_pairs_selection} top-right. By visually comparing both top pictures, one can qualitatively say the algorithm has correctly selected feasible pairs from the candidates. Fig. \ref{fig:pole_pairs_selection} bottom shows point clouds after this calibration step. Note how pole-like objects match (red, orange, purple points from trees), and how the street widens as the sensors are now correctly centered around the vehicle.

\begin{figure}[b]
	\centering
	\medskip
	\includegraphics[width=\linewidth, trim={0cm 6cm 6cm 6cm}, clip]{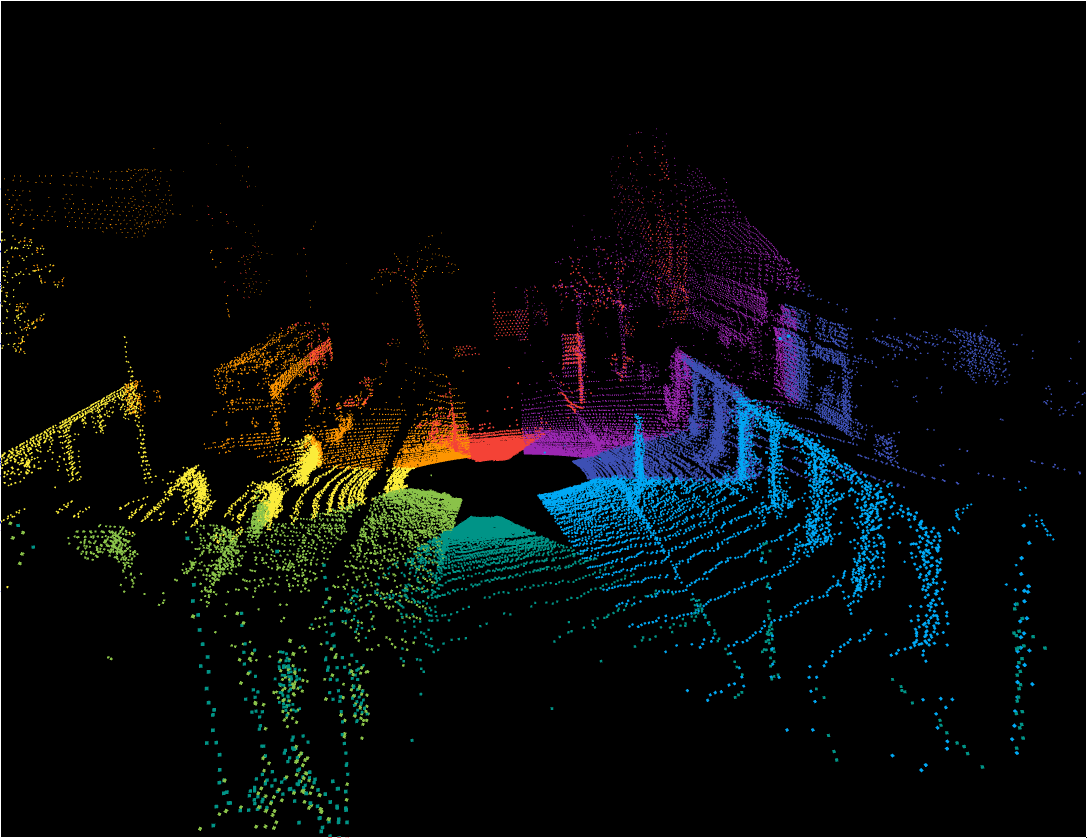}
	\caption[]{\small Resulting alignment of point clouds after joint cross-sensor calibration (algorithm presented in section \ref{sec:cross_estimation}).}
	\label{fig:cross_calibration}
\end{figure}

Finally, in the last step of the calibration algorithm (section \ref{sec:cross_estimation}), we estimate all calibration parameters and produce an overall point cloud, whose alignment quality is visualized in Fig. \ref{fig:cross_calibration}. See e.g. how the height of the red point cloud is now correct. A comparison between the estimated calibrations and the ground truth ones, on a set of 50 experiments (that use small subsets of information from long trips recorded in an urban scenario), produces the following results: The total translation error has a mean of 12.37 cm and standard deviation of 0.99 cm, and the total orientation error has a mean of 0.91 deg and standard deviation of 0.08 deg. For this experiment, input features (pole-like objects) were computed similarly as suggested in \cite{AlexPaperOnPolesDetection}. Solving the optimization problems involved within the offline version of the calibration algorithm takes typically 15 seconds. The most computationally demanding part is solving the MIP, which takes Gurobi \cite{gurobi} around 10 seconds. However, given that the MIP only needs to be solved once and in an offline fashion, the CBC solver \cite{john_forrest_2018_1317566} can also be used to this end.

Having shown an evaluation of our algorithm using data from a real scenario and seen that it produces very good calibration results (comparable in accuracy to other approaches, but with the additional benefits of being able to run in a dynamic environment, no need for dedicated targets, or non-degenerate trajectories), we evaluate its performance using simulation data for which ground truth is available.

\begin{figure}[t]
	\centering
	\medskip
	\includegraphics[width=0.49\linewidth, trim={0cm 1.5cm 0cm 3cm}, clip]{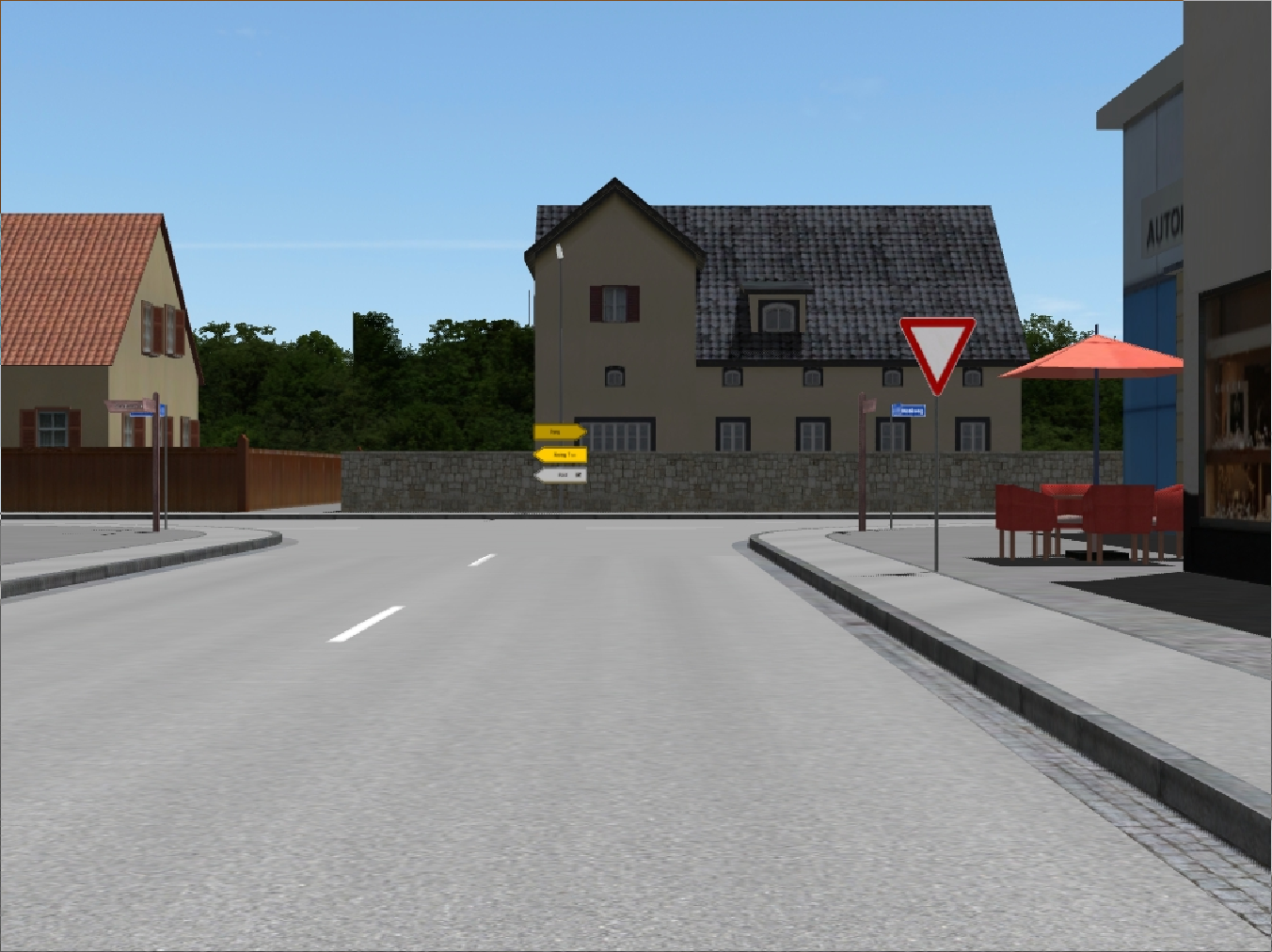}
	\includegraphics[width=0.49\linewidth, trim={3cm 1cm 3cm 7.5cm}, clip]{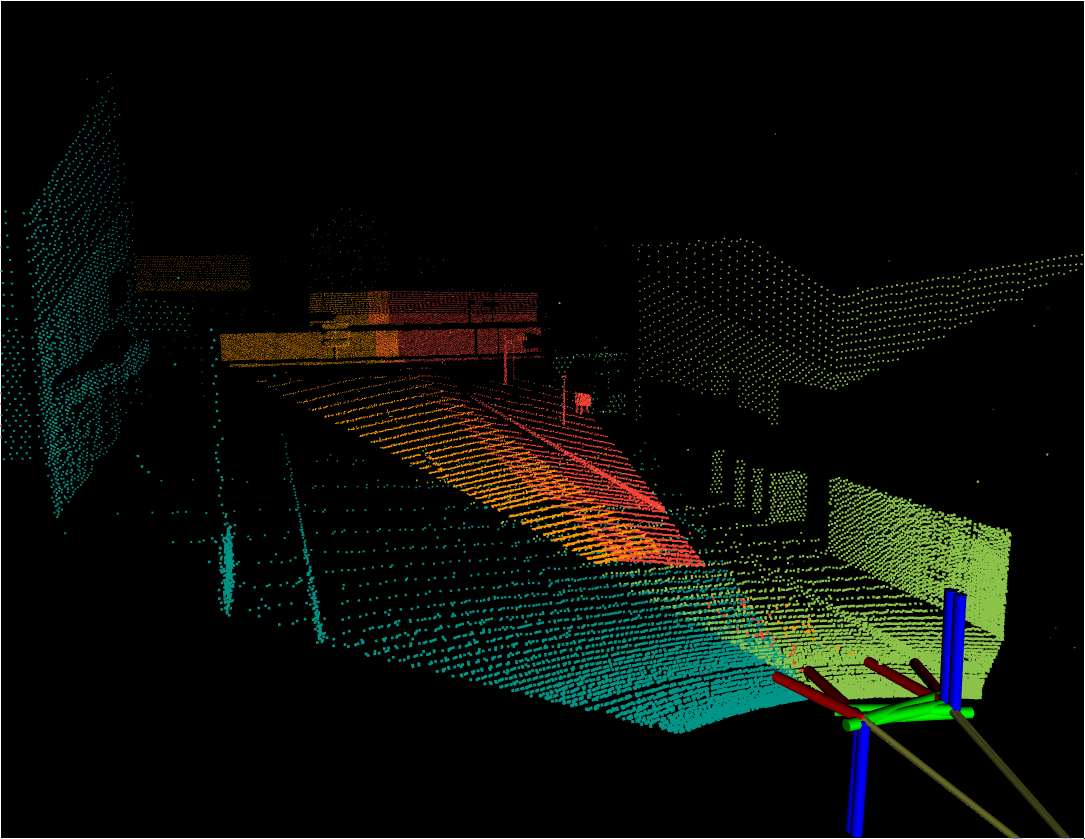}
	\includegraphics[width=\linewidth, trim={0cm 0cm 0cm 0cm}, clip]{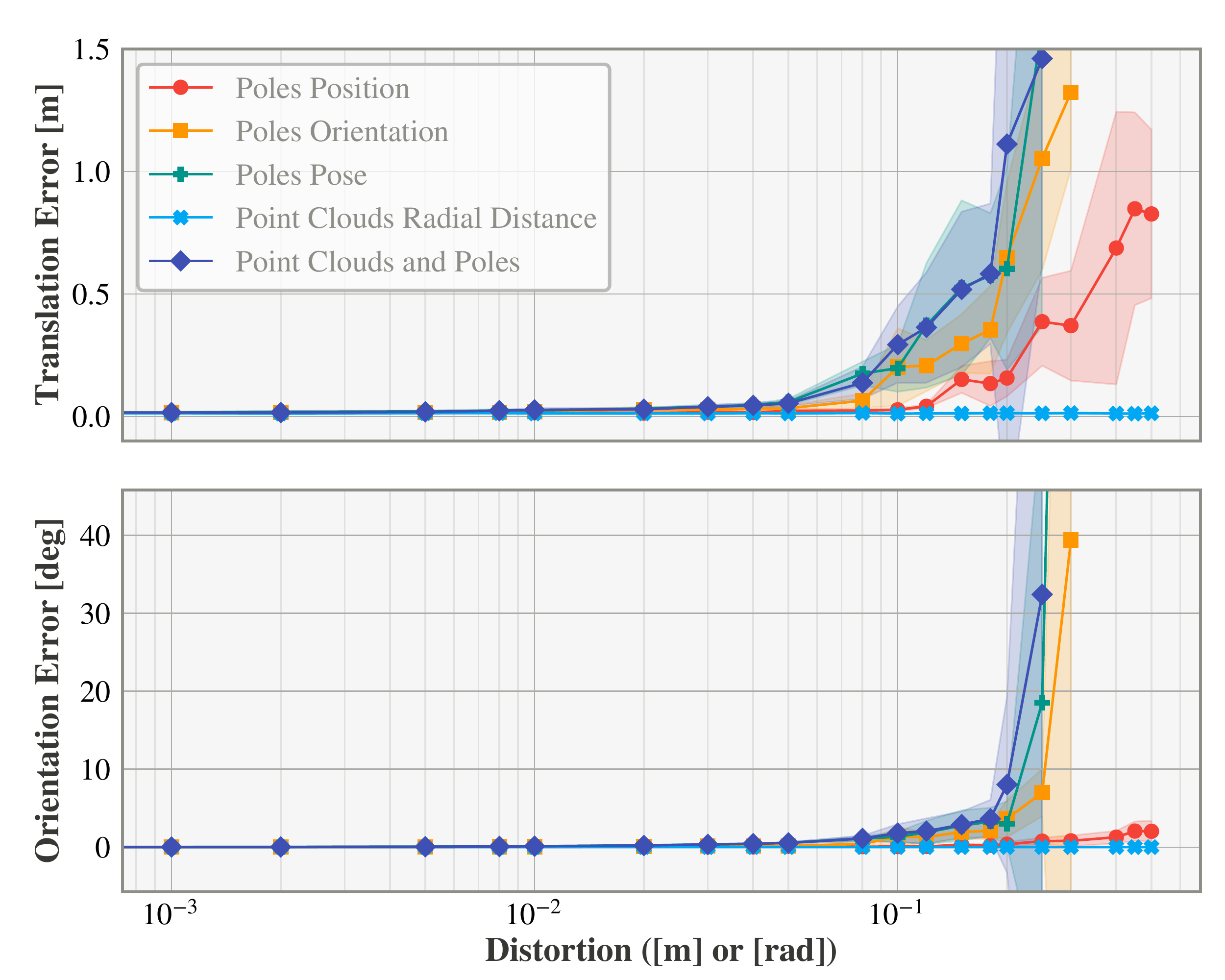}
	\caption[]{\small \textit{Top}: Picture of the urban simulation scenario and vehicle setup with 4 sensors used for evaluation. \textit{Bottom}: Per-sensor translation and orientation errors (mean and standard deviation) of the calibration algorithm (presented in  section \ref{sec:approach_offline_calibration}) as a function of the amount of distortion of the inputs. }
	\label{fig:audi_simulation_offline}
\end{figure}

For this experiment, we use a different sensor setup, where the vehicle is equipped with four sensors with different field of view angles, as shown in Fig. \ref{fig:audi_simulation_offline} top row, and the scenario is an urban environment as in the previous example. The inputs to our algorithm are pole detections as well as ground points. Thus, we evaluate the accuracy and robustness of our calibration algorithm by applying different perturbations to these inputs. In the bottom of Fig. \ref{fig:audi_simulation_offline}, we show the resulting calibration accuracy for different distortion types and magnitude (statistics for each point are computed based on 10 experiments with randomly sampled distortions). For instance, the distortion type "Poles Position" distorts just the XY location of the poles (as perturbing its Z component in vehicle coordinates would not change the distance between a pair of poles), the type "Poles Orientation" rotates the pole base and top coordinates using a random orientation, and "Poles Pose" distorts poles using a random translation and orientation. The distortion type "Point Clouds Radial Distance" adds noise to the input ground points along the radial direction, and the final distortion type "Point Clouds and Poles" perturbs all components. The amount of distortion for each distorted component of the poles and ground points comes from a uniform distribution whose minimum and maximum absolute values correspond to the plot x-values. 
The units of the distortion amount depend on the component being distorted, i.e. for poles position and point clouds radial distance it is measured in meters, and for poles orientation in radians. In the results from Fig. \ref{fig:audi_simulation_offline}, we can see that as expected the larger the distortion amount, the larger the calibration error. Note that for most of the distortion types, good calibration values can only be robustly recovered up to distortion amounts of 0.1 units, except for the "Poles Position" distortion type where it can be done up to distortion amounts of 0.2 meters. The error of our static objects detector lies in this range so that improving the calibration accuracy is infeasible given the object accuracy.

\subsection{Evaluation of Online Calibration Approach}

In this section, we show the calibration results when applying the algorithm online (section \ref{sec:approach_online_calibration}). To this end, we use simulation data (where known distortions can be applied to the sensor mounting poses) as well as real data (where the initial calibration values can be slightly distorted).

\begin{figure}[b]
	\medskip
	\;\;\;\;\includegraphics[width=0.80\linewidth, trim={0cm 0cm 3cm 0.5cm}, clip]{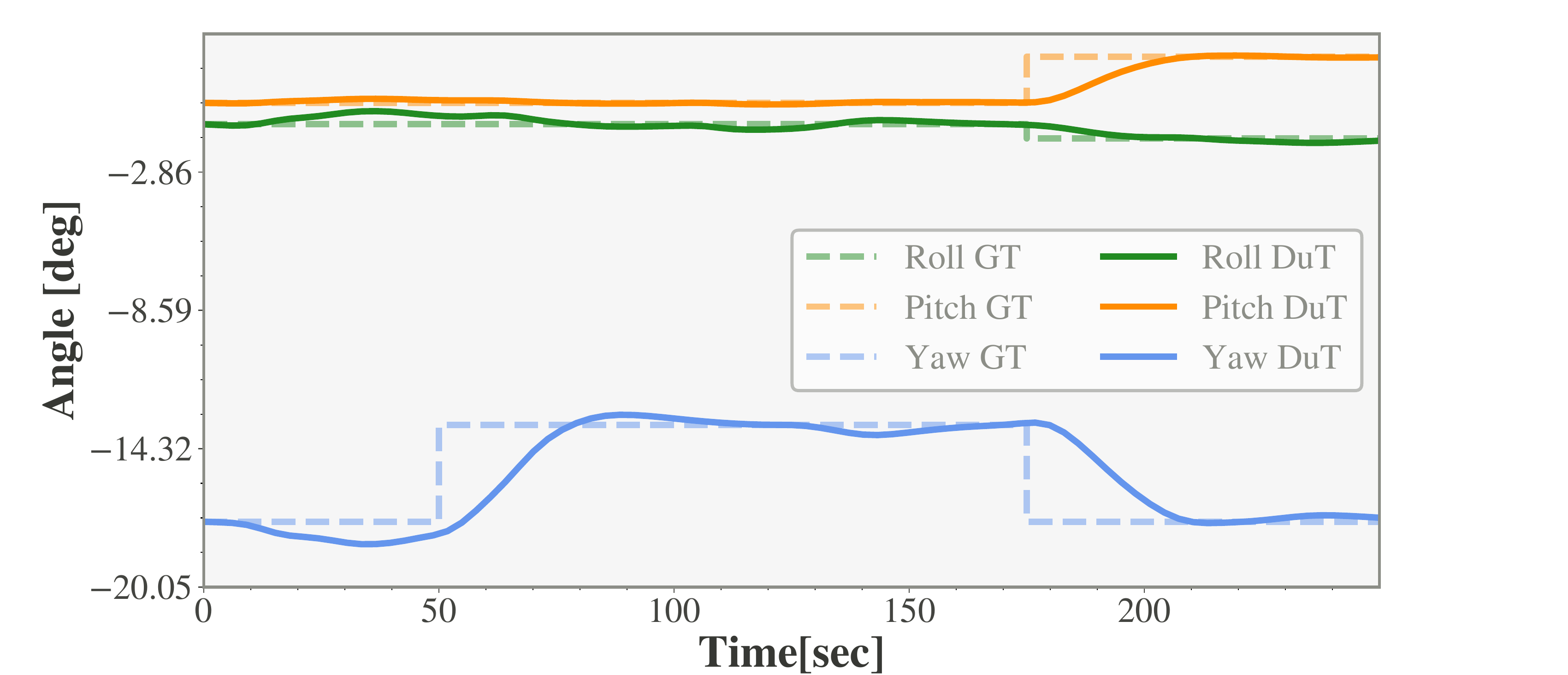}
	\caption[]{\small Online calibration in simulation scenario. One of the sensors' mounting pose is perturbed twice at 50 and 175 [sec]. GT denotes ground truth and DuT the estimated calibration values.}
	\label{fig:audi_simulation_online}
	\centering
	\includegraphics[width=0.92\linewidth, trim={0cm 0cm 0cm 0cm}, clip]{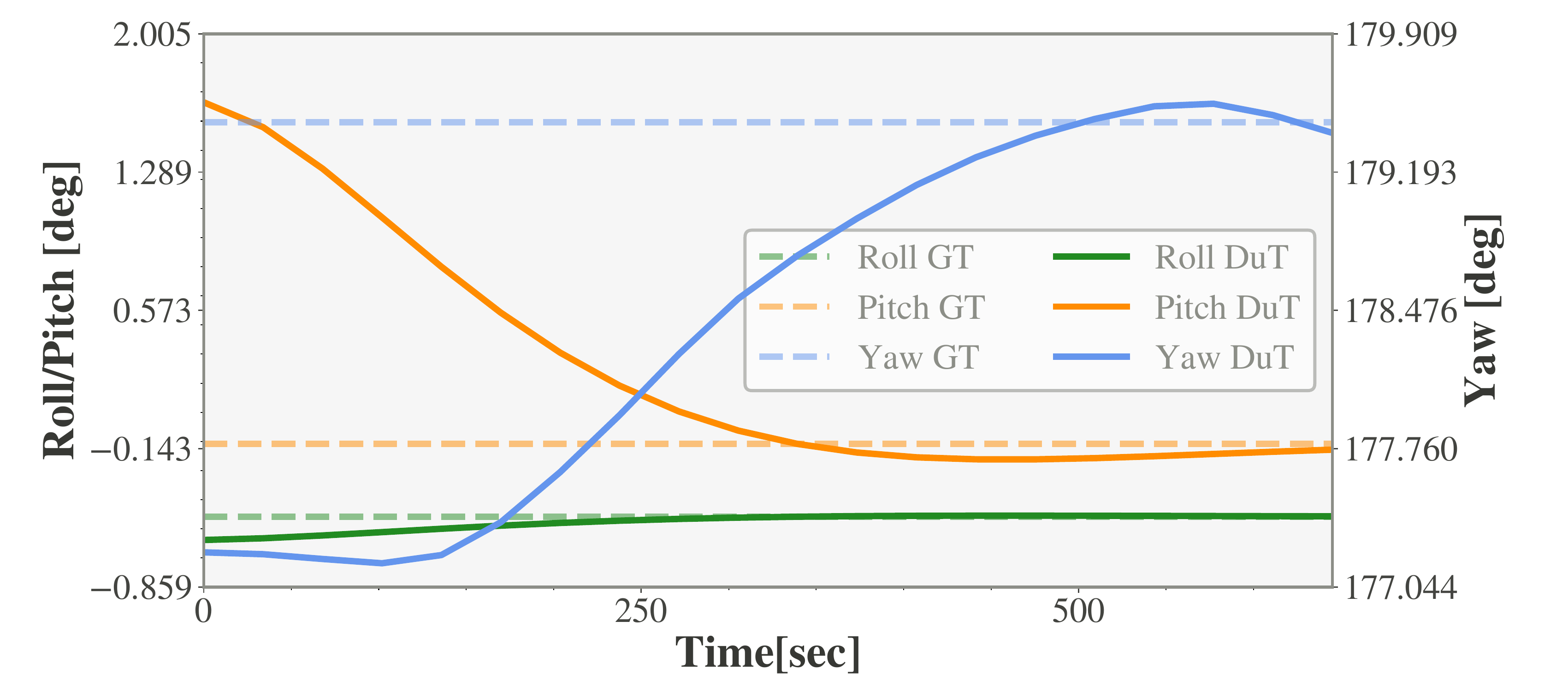}
	\caption[]{\small Online calibration in real scenario. Calibration estimates of orientation angles converging towards ground truth values.  }
	\label{fig:vclass_simulation_online}
\end{figure}

Fig. \ref{fig:audi_simulation_online} shows results from a simulation scenario (vehicle with 4 sensors), where the mounting pose of one of the sensors is changed at two points in time (50 and 175 sec). We can see how the optimized calibration values closely track these changes in an online fashion. In Fig. \ref{fig:vclass_simulation_online} we show as well online calibration results, but in this case for the urban scenario using real data (vehicle with 8 sensors), where we initialize the experiment with calibration values away from the ground truth ones (for all sensors). In a similar way, we can see that also in this case the calibration values converge to the ground truth ones over time. Similar behavior can be seen for the rest of the sensors (not shown). The online version of the calibration algorithm takes typically 30 ms.

To conclude, we note some limitations of our approach. First of all, our approach is targeted to sensors providing 3D information such as LiDAR, a usability for other sensors such as monocular cameras was not further examined here. Second, no direct estimation of the relative position between IMU and LiDAR sensors is provided, but can be inferred from the location of the IMU in $\mathbf{K}_{\mathrm{V}}$, which can be easily measured by other methods.

\section{Conclusion} \label{sec:conclusion}

We have presented an efficient algorithm for the calibration of sensors rigidly mounted in an autonomous vehicle that exploits proprioceptive/perceptual information as well as time/space information. Finally, we have shown experimental evidence on simulation and real data (urban environment) of how our approach is able to find accurate calibration values.







\bibliography{references}
\bibliographystyle{ieeetr}

\end{document}